\documentclass[conference]{IEEEtran}

\ifdefined\pdfminorversion \pdfminorversion=7 \fi  

\usepackage{times}
\usepackage{soul}
\usepackage{url}
\usepackage[hidelinks]{hyperref}
\usepackage[utf8]{inputenc}
\usepackage{graphicx}
\usepackage{amsmath}
\usepackage{amsthm}
\usepackage{booktabs}
\usepackage{algorithm}
\usepackage{algorithmic}
\usepackage{multirow}
\usepackage[table]{xcolor}
\usepackage{bbm}
\usepackage{amssymb}
\usepackage{makecell}
\usepackage{colortbl}
\usepackage{array}
\usepackage[final]{microtype}
\usepackage{xurl}
\usepackage{cite}
\usepackage{threeparttable}
\usepackage{float}
\begin{document}

\title{CogAdapt: Adapting Clinical ECG Foundation Models for Wearable Cognitive Load Assessment}

\author{
\IEEEauthorblockN{
Amir Mousavi\textsuperscript{1},
Erfan Nourbakhsh\textsuperscript{1},
Mohammad Sadegh Sirjani\textsuperscript{1},
Rocky Slavin\textsuperscript{1},
Mimi Xie\textsuperscript{1}\\
Leslie Neely\textsuperscript{3},
John Davis\textsuperscript{2},
John Quarles\textsuperscript{1}
}

\IEEEauthorblockA{
\textsuperscript{1}Department of Computer Science, College of AI, Cyber and Computing,\\
The University of Texas at San Antonio\\
\{seyedamir.mousavi, erfan.nourbakhsh, mohammadsadegh.sirjani\}@utsa.edu\\
\{ rocky.slavin, mimi.xie, john.quarles\}@utsa.edu
}

\IEEEauthorblockA{
\textsuperscript{2}Department of Educational Psychology, College of Education and Human Development,\\
The University of Texas at San Antonio\\
john.davis2@utsa.edu
}

\IEEEauthorblockA{
\textsuperscript{3}Department of Neuroscience, Developmental and Regenerative Biology, College of Sciences,\\
The University of Texas at San Antonio\\
leslie.neely@utsa.edu
}
}

\maketitle

\begin{abstract}
Assessing cognitive load continuously and at low latency would help adaptive human-computer interaction, but it remains hard because labeled data are scarce and models generalize poorly across subjects. Recent ECG foundation models, pre-trained on millions of clinical diagnostic ECG recordings, yet they do not apply directly to wearable devices when the sensor configuration and the task both differ. We present CogAdapt, a framework that adapts a clinical ECG foundation model to wearable cognitive load assessment. CogAdapt has two parts. LeadBridge is a learnable adapter that maps 3-lead wearable signals to a 12-lead-compatible representation. ProFine is a progressive fine-tuning strategy that unfreezes encoder layers in stages while limiting representational drift in the pre-trained model. On two public datasets (CLARE and CL-Drive) under leave-one-subject-out cross-validation, CogAdapt reaches macro-F1 of 0.626 and 0.768, improving over from-scratch baselines by 11.2 and 16.1 percentage points. The results show that a clinical ECG pretraining can support subject-independent cognitive load assessment from wearable sensors.
\end{abstract}


\section{Introduction}
\label{sec:introduction}

Cognitive load, as formalized by Sweller's Cognitive Load Theory~\cite{sweller1988cognitive}, refers to the demands placed on working memory during information processing. Accurate, low-latency assessment of cognitive load would benefit adaptive human-computer interaction, but it remains hard with physiological signals. On a cognitive load real-time assessment dataset (CLARE) , the best reported ECG-based model was CNN, which showed a drop from 78.4\% accuracy under $k$-fold evaluation to 63.3\% under subject-independent evaluation~\cite{bhatti2024clare}. Models trained from scratch on small wearable cognitive load datasets can learn stable differences between participants instead of mental effort dependent physiological changes. For ECG, these differences include baseline heart rate, signal amplitude, electrode placement effects, and individual autonomic reactivity. 

This generalization failure has a structural cause. Cognitive load datasets are small. CLARE contains 20 usable subjects; CL-Drive~\cite{angkan2024multimodal} contains 21 participants. Few public cognitive-load datasets support real-time assessment. Both provide fine-grained 10-second labels, which is a good fit for window-level evaluation, but their subject pools are insufficient for complex models to learn cross-subject representations. Under leave-one-subject-out (LOSO), the model trains on at most 20 subjects and tests on a new subject. ECG amplitude, heart-rate baseline, and autonomic reactivity vary across people. This makes the task a cross-subject transfer problem, not only a classification problem.  

Electrocardiography (ECG) is a practical signal for cognitive load monitoring because mental effort modulates the autonomic nervous system (ANS), producing measurable changes in heart rate variability and cardiac morphology~\cite{Hughes2019,Ayres2021TheVO,Raza2024SystematicRO}. Increased cognitive demand activates the prefrontal cortex, which modulates the hypothalamic-autonomic pathway and reduces vagal tone, leading to decreased parasympathetic influence on heart rate~\cite{thayer2009neurovisceral}. Clinical ECG foundation models pretrained on millions of 12-lead recordings of cardiac representations that transfer across diagnostic tasks~\cite{Gu_2025,mckeen2024ecgfm}. However, two barriers prevent direct transfer to wearable cognitive load assessment. (1)~A \emph{sensor gap} --- clinical models expect 12-lead ECG input, but wearable devices typically provide only 1-3 leads. (2)~A \emph{task gap} --- clinical models are pretrained for pathology detection, not the subtle autonomic modulation associated with cognitive effort. Prior work on wearable-to-clinical ECG adaptation~\cite{Hassannia2025ANN,BioXBridge} addresses the sensor gap in isolation but does not target cognitive load classification or resolve the task gap.

\begin{figure}[t]
    \centering
    \includegraphics[width=\linewidth]{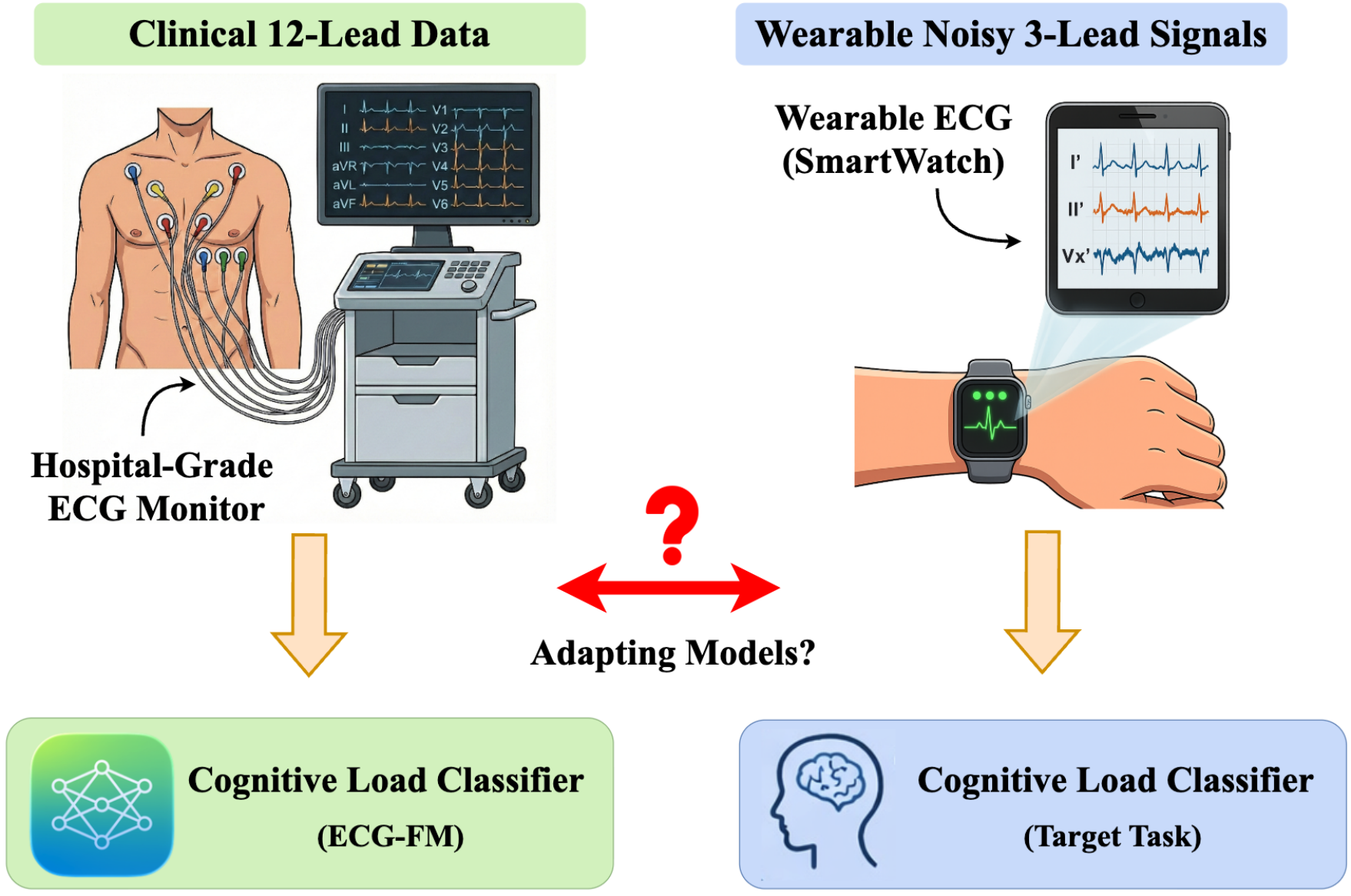}
    \caption{The core challenge: pretrained foundation models expect 12-lead clinical ECG, but wearable cognitive load datasets provide only 3-lead recordings.}
    \label{fig:intro}
\end{figure}

We present \textbf{CogAdapt}, a framework that closes both gaps (Figure~\ref{fig:intro}). \textbf{LeadBridge} is a lightweight convolutional adapter pretrained on PTB-XL~\cite{Wagner2020PTBXL} that maps 3-lead wearable ECG to a 12-lead-compatible representation. \textbf{ProFine} is a tiered fine-tuning strategy that incrementally unfreezes the pretrained encoder, adapting task-specific representations while limiting representational drift via bucketed layer-wise learning rates.

Our main contributions are:
\begin{itemize}
  \item We introduce \textbf{LeadBridge}, a lightweight adapter that learns to map 3-lead wearable ECG to an anatomically consistent 12-lead form, pretrained on PTB-XL and further optimized end-to-end with the classification objective.
  \item We design \textbf{ProFine}, a tiered fine-tuning procedure that gradually adapts the foundation encoder to cognitive load labels while limiting representational drift via bucketed layer-wise learning rates.
  \item On CLARE and CL-Drive under LOSO evaluation, CogAdapt reaches macro-F1 of \textbf{0.626} and \textbf{0.768}, gains of $+0.112$ and $+0.161$ over the strongest from-scratch baseline (ECG-LightCNN), both significant by a Wilcoxon signed-rank test ($p{<}0.05$). We also report cross-dataset transfer and inference latency to gauge deployment feasibility.
\end{itemize}

\section{Related Work}
\label{sec:related}

\subsection{Cognitive Load Datasets and LOSO Generalization}

Cognitive load datasets differ mainly in label granularity. Earlier datasets such as Gjoreski et al.~\cite{gjoreski2020cogload} and CLAS~\cite{markova2019clas} report one rating after each task or block, which is too coarse for window-level classifiers or it introduces memory recency bias in labeling. CLARE~\cite{bhatti2024clare} and CL-Drive~\cite{angkan2024multimodal} instead provide a load rating every 10 seconds, during MATB-II multitasking and simulated driving. Denser labels do not solve generalization. With only about 20 subjects per set, the CLARE CNN that scores well under $k$-fold collapses under subject-independent evaluation~\cite{bhatti2024clare}, fitting subject-specific ECG rather than load-invariant features. Closing this subject-independent gap is what we target.

\subsection{ECG-Based Cognitive Load Modeling}

Physiological models of cognitive load use either hand-crafted heart-rate-variability (HRV) features or end-to-end deep networks. HRV features are interpretable and cheap, but individual baseline variability limits their sensitivity~\cite{Raza2024SystematicRO,Hughes2019}. Even a richer 35-feature time- and frequency-domain HRV/ECG set (our HRV-RF baseline) reaches only F1 0.444 on CLARE and 0.533 on CL-Drive under LOSO. CNNs and transformers on raw ECG can beat HRV features in-distribution, yet they overfit on small sets~\cite{bhatti2024clare,Ayres2021TheVO}. With roughly 20 training subjects, any from-scratch model sits in a low-data regime for cross-subject transfer. This motivates transfer from a larger source.

\subsection{ECG Foundation Models}

Self-supervised pretraining on large clinical ECG corpora yields foundation models that cut labeled-data needs for cardiac tasks~\cite{Gu_2025,mckeen2024ecgfm,han2025systematicreviewfoundationmodels}. ECG-FM reaches AUROC 0.996 on atrial fibrillation with little labeled data~\cite{mckeen2024ecgfm}. CLEF weights contrastive pairs by clinical risk to improve single-lead wearable classification~\cite{clef2025}, and NormWear extends pretraining to multimodal wearable sensing across ECG, PPG, EEG, and GSR~\cite{normwear2025}. Two gaps block direct use for wearable cognitive load. First, ECG-FM and ECGFounder expect 12-lead input, while wearables give 2--3 leads, so zero-padding or masking breaks the spatial structure the encoder learned in pretraining. Second, these models are pretrained for pathology, whereas cognitive load appears as subtle autonomic change. Neither CLEF nor NormWear has been evaluated on cognitive load. CogAdapt targets both gaps.

\subsection{Clinical-to-Wearable ECG Adaptation}

Reconstructing 12 leads from a reduced set has used fixed transforms, linear regression, and learned mappings. The Dower inverse transform derives precordial leads from Frank XYZ leads with fixed coefficients~\cite{dower1980deriving}, but it assumes standard electrode geometry, which fails for consumer wearables~\cite{FEILD2008466,VOZDA201523}. Linear regression corrects for some patient variation yet stays static~\cite{obianom2025reconstruction}. Learned mappings adapt better: Hassannia and Sameni~\cite{Hassannia2025ANN} show a shared and lead-specific network beats regression, and BioX-Bridge~\cite{BioXBridge} aligns biosignal foundation models through a small bridge network. None of these target cognitive load or the task gap. CogAdapt differs on two points. LeadBridge maps leads for foundation-model compatibility, optimizing the encoder input format rather than reconstruction fidelity. ProFine then adapts the encoder to the load task. Together they form one pipeline from 12-lead clinical pretraining to 3-lead wearable cognitive load classification.

\section{Proposed Methodology}
\label{sec:methodology}

\subsection{Framework Overview}
\label{sec:overview}

\begin{figure*}[t]
    \centering
    \vspace{-1em}
    \includegraphics[width=0.82\textwidth]{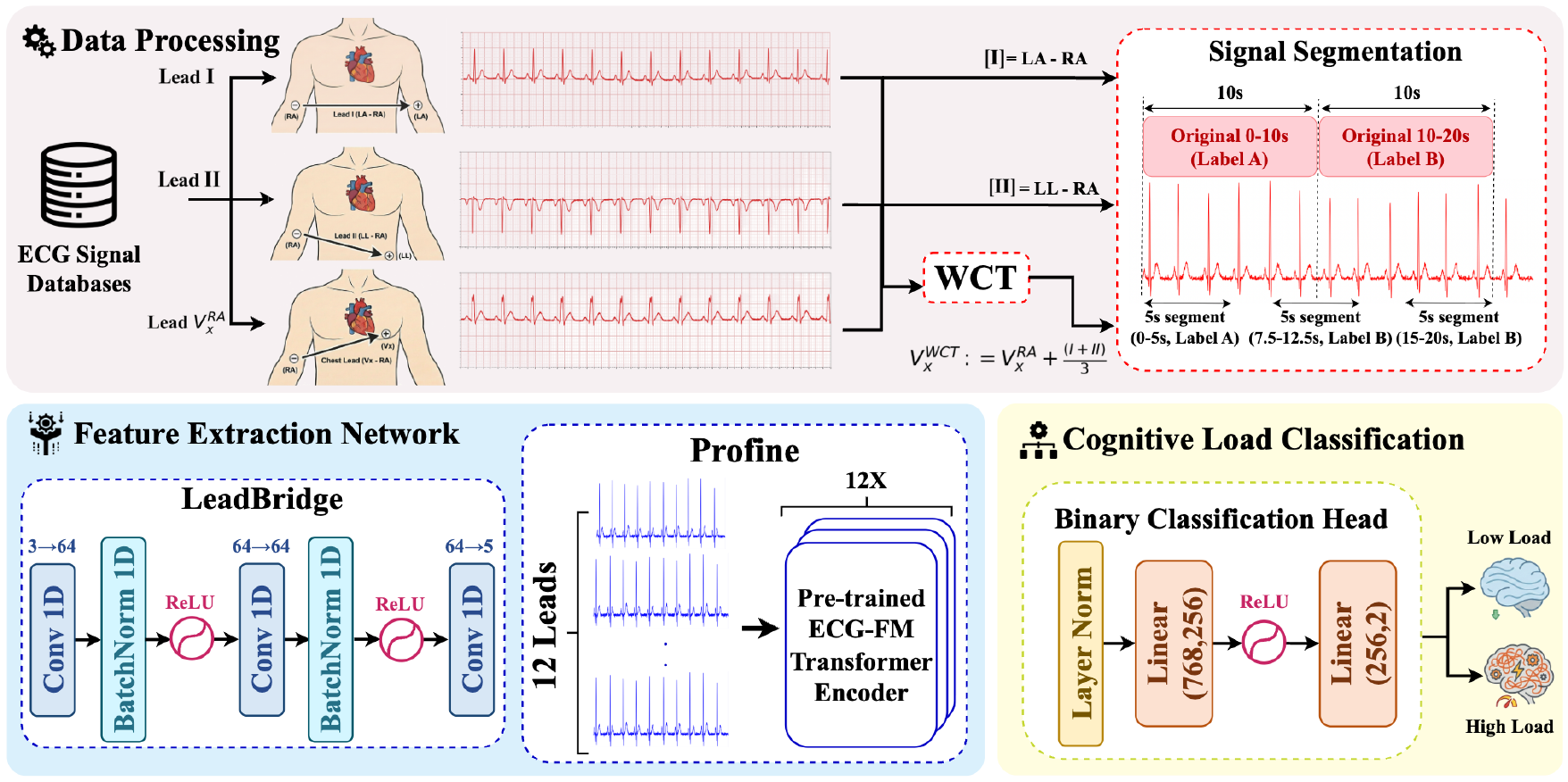}
    \caption{The CogAdapt pipeline: LeadBridge ($3{\to}12$ leads), the pretrained ECG-FM encoder, and ProFine fine-tuning for cognitive-load classification.}
    \label{fig:architecture}
\end{figure*}

Figure~\ref{fig:architecture} shows the CogAdapt framework. CogAdapt adapts a clinical ECG foundation model to wearable cognitive-load classification. The framework addresses two gaps. The first gap is the sensor-configuration mismatch between 3-lead wearable ECG and 12-lead clinical ECG. The second gap is the task mismatch between clinical ECG pretraining and cognitive-load classification.

CogAdapt has four components. First, the data processing pipeline filters the raw wearable ECG, applies Wilson Central Terminal (WCT) re-referencing, extracts windows, and normalizes each window. Second, LeadBridge maps each 3-lead wearable window into a 12-lead-compatible representation. Third, ECG-FM encodes the adapted ECG window. Fourth, ProFine controls how much of the pretrained encoder is updated during cognitive-load training.

Together, LeadBridge and ProFine let a clinical 12-lead ECG encoder process 3-lead wearable ECG without changing its input layer.

\subsection{Problem Formulation}
\label{sec:problem}

We formulate cognitive-load assessment as supervised time-series classification under domain shift. Let $\mathbf{x}\in\mathbb{R}^{C\times T}$ denote an ECG window with $C$ leads and $T$ samples. Let $y\in\{0,1\}$ denote the cognitive-load label, where $0$ is low load and $1$ is high load.

The source model is a pretrained ECG encoder $E_\theta$. It expects clinical 12-lead ECG input, so $C_S=12$. The target datasets provide wearable 3-lead ECG, so $C_T=3$. Each target example is therefore $(\mathbf{x}_i,y_i)$, where $\mathbf{x}_i\in\mathbb{R}^{3\times T}$.

Direct transfer faces the two gaps of Section~\ref{sec:overview}: the encoder expects 12 leads but the wearable provides three, and it was pretrained for clinical tasks rather than cognitive load. Naive lead expansion can break the lead structure learned in pretraining.

CogAdapt learns a prediction function
\begin{equation}
\Phi(\mathbf{x}) = g_\psi\big(\text{Pool}(E_\theta(A_\phi(\mathbf{x})))\big),
\end{equation}
where $A_\phi$ is LeadBridge, $E_\theta$ is the pretrained ECG encoder, and $g_\psi$ is the classification head. LeadBridge maps $\mathbf{x}$ from 3 leads to 12 leads. ProFine controls which parts of $E_\theta$ are updated during training.

\subsection{Wearable ECG Data Processing}
\label{sec:input}

We resample each recording to 500~Hz and band-pass filter it (0.5--40~Hz). We drop samples with all leads missing. We zero-fill sporadic missing values after filtering. We set constant leads to zero within a window so normalization stays stable. The wearable provides Lead~I, Lead~II, and one right-leg-referenced chest lead $V_x^{\mathrm{RL}}$. Clinical chest leads use Wilson's Central Terminal (WCT)~\cite{moeinzadeh2018modern}, so, following the CLARE convention~\cite{bhatti2024clare} and the approximation $\mathrm{RL}\approx\mathrm{RA}$, we re-reference to
\begin{equation}
V_x^{\mathrm{WCT}} = V_x^{\mathrm{RL}} + \tfrac{I + II}{3},
\end{equation}
which gives the 3-lead input $\mathbf{x}_{\mathrm{ref}} = [I, II, V_x^{\mathrm{WCT}}]^\top$. Its chest lead now uses the clinical WCT reference, matching the 12-lead recordings ECG-FM was pretrained on. LeadBridge later expands these three leads to the 12 the encoder requires.

We segment 5-second windows and label each by the rating interval that contains its midpoint, so the label is not tied to a window's start or end alone. The 5-second span covers the 2--5\,s autonomic response latency~\cite{Hughes2019}. Training uses 50\% overlap and evaluation uses non-overlapping windows. Overlap is applied only within a fold's training subjects, so no held-out-subject window enters training and no training window overlaps a test window. We z-score each window per lead, which removes electrode- and subject-specific amplitude and, computed within the window, leaks no test-subject statistics.

\subsection{LeadBridge: Learnable $3\!\rightarrow\!12$ Lead Adaptation}
\label{sec:adapter}

LeadBridge maps 3-lead wearable ECG to a 12-lead-compatible representation: $A_\phi: \mathbb{R}^{3\times T}\rightarrow\mathbb{R}^{12\times T}$. The goal is not to replace a clinical 12-lead ECG recording, but to produce an input format that the pretrained encoder can process while preserving lead relationships.

CLARE and CL-Drive do not provide paired 3-lead and 12-lead recordings. We therefore pretrain LeadBridge on normal sinus rhythm recordings from PTB-XL~\cite{Wagner2020PTBXL} (${\approx}9{,}500$ of 21,837 after quality filtering), simulating the wearable 3-lead configuration from the clinical leads. The pretraining objective is the mean-squared error on the five learned precordial leads V2--V6; the analytic limb and augmented leads carry no learned loss. We optimise with AdamW (learning rate $0.003$, weight decay $0.02$, batch size 128) for up to 400 epochs, with early stopping (patience 20). Following PTB-XL's predefined stratification folds~\cite{Wagner2020PTBXL}, we train on folds 1--8 (${\approx}7{,}600$ recordings), validate on fold 9 (${\approx}950$), and hold out fold 10 (${\approx}950$) as the test set for the reconstruction evaluation (Section~\ref{sec:lead_reconstruction}); subjects are disjoint across folds.

\subsubsection{Adapter Architecture}

LeadBridge uses a two-stage design. The first stage forms seven leads without learning their shape. LeadI, LeadII, and the chest lead $V_x$ (clinically $V_1$) pass through, while III, aVR, aVL, and aVF follow from Einthoven's and Goldberger's relations. A learnable per-lead scale-and-bias correction ($14$ parameters) then matches their amplitudes:
\begin{align}
\mathbf{III} &= \mathrm{II} - \mathrm{I},\quad
\mathrm{aVR} = -(\mathrm{I}+\mathrm{II})/2,\\
\mathrm{aVL} &= \mathrm{I}-\mathrm{II}/2,\quad
\mathrm{aVF} = \mathrm{II}-\mathrm{I}/2,
\end{align}
giving $[\mathrm{I},\mathrm{II},\mathrm{III},\mathrm{aVR},\mathrm{aVL},\mathrm{aVF},V_x]^\top$.
The second stage learns the five precordial leads V2--V6 from the 3-lead input via a point-wise Conv1D network (kernel size~1, no temporal mixing):
\begin{equation}
\hat{\mathbf{V}}_{2:6} = W_3\,\sigma\!\left(\mathrm{BN}\left(W_2\,\sigma\!\left(\mathrm{BN}(W_1\mathbf{x})\right)\right)\right),
\label{eq:adapter}
\end{equation}
where $W_1\!\in\!\mathbb{R}^{64\times3}$, $W_2\!\in\!\mathbb{R}^{64\times64}$, $W_3\!\in\!\mathbb{R}^{5\times64}$. The full 12-lead output is $[\text{7 limb/augmented}, \hat{\mathbf{V}}_{2:6}]^\top$. LeadBridge has 4,883 trainable parameters in total.

This design is lightweight but more expressive than fixed transforms: the limb and augmented leads are computed analytically, so only the harder-to-estimate precordial leads are learned. During cognitive-load training (Scenarios A--C), LeadBridge is updated end-to-end with the classification head, and with the encoder in B and~C, on the CLARE/CL-Drive training folds. Its learning rate is set below the head's, which preserves the learned lead structure while allowing task-oriented adaptation; this shifts LeadBridge from reconstruction-optimal to task-adapted.

\subsection{ProFine: Progressive Fine-Tuning}
\label{sec:finetuning}

ProFine adapts ECG-FM to cognitive-load classification through three training scenarios that differ in how much of the pretrained encoder is updated. It follows the progressive-unfreezing tradition of ULMFiT~\cite{howard2018ulmfit}, here with layer-wise learning-rate buckets tuned for the ECG-FM encoder; parameter-efficient alternatives such as LoRA~\cite{hu2022lora} are complementary and left to future work. Let the encoder contain $L$ layers, denoted $\{\ell_1,\dots,\ell_L\}$.

\begin{figure}[t]
    \centering
    \includegraphics[width=0.78\linewidth]{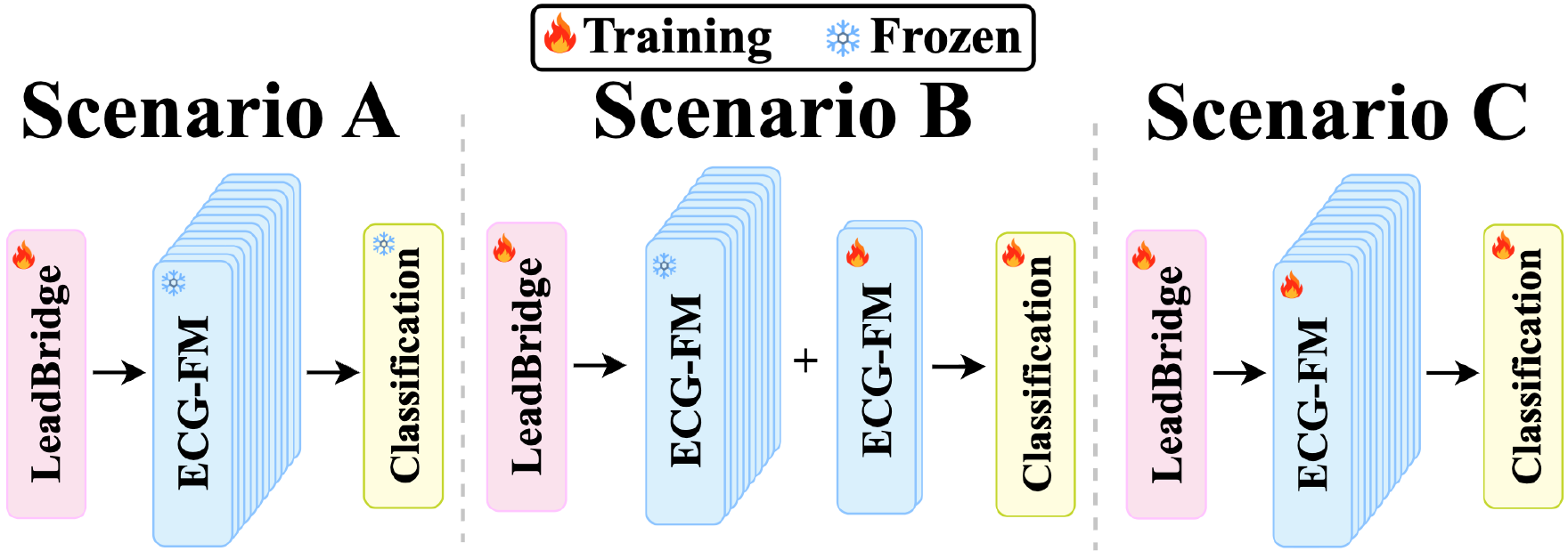}
    \caption{Progressive fine-tuning scenarios. Scenario~A freezes the encoder. Scenario~B unfreezes top layers. Scenario~C unfreezes all layers.}
    \label{fig:stages}
\end{figure}

\subsubsection{Scenario A: Frozen Encoder}

Scenario~A freezes the full ECG-FM encoder and trains only LeadBridge and the classification head: $\Theta_{\mathrm{train}}^A=\{\phi,\psi\}$. This tests whether fixed clinical ECG features are enough after lead adaptation.

\subsubsection{Scenario B: Partial Fine-Tuning}

Scenario~B unfreezes the top $K=2$ encoder layers: $\Theta_{\mathrm{train}}^B=\{\phi,\psi\}\cup\{\ell_j\mid j>L-K\}$. This adapts high-level representations while keeping lower ECG morphology layers stable.

\subsubsection{Scenario C: Full Fine-Tuning}

Scenario~C updates all parameters: $\Theta_{\mathrm{train}}^C=\{\phi,\psi,\ell_1,\dots,\ell_L\}$. We use bucketed layer-wise learning rates: the top, middle, and bottom thirds of the encoder receive progressively smaller updates. This preserves general ECG features while still adapting to cognitive-load labels.

\subsection{Classification Head and Training Objective}
\label{sec:objective}

The classification head maps pooled ECG-FM features to binary cognitive-load predictions. We mean-pool the encoder output over time, apply LayerNorm, and pass the result through a two-layer MLP: a linear layer to 256 units, a ReLU, dropout ($p{=}0.2$), and a final linear layer to two logits. We use this lightweight two-layer MLP, rather than the single linear probe used in ECG-FM pretraining, to permit a nonlinear decision boundary over the pooled ECG-FM representation. The encoder hidden dimension is 768.

We use ECG-FM's publicly available pretrained checkpoint trained on MIMIC-IV-ECG and PhysioNet 2021~\cite{mckeen2024ecgfm}. No additional clinical data collection was required.

All three scenarios train with cross-entropy loss. The labels lean toward high load: the high-to-low ratio is about $2.0{:}1$ on CLARE (66.7\% of 38{,}064 windows) and $1.4{:}1$ on CL-Drive (58.5\% of 19{,}782). Under LOSO the per-subject imbalance is far wider, from roughly balanced to beyond $50{:}1$ in either direction, and one CLARE subject is single-class.

We do not reweight the loss. We select checkpoints with early stopping (patience~10), using AUROC under stratified K-fold and macro-F1 under LOSO (Section~\ref{sec:experimental_setup}).

All scenarios use AdamW~\cite{loshchilov2017decoupled} at batch size 32, weight decay $10^{-4}$, and a cosine learning rate schedule with 10\% warmup upto 100 epochs. The head and adapter learning rates fall as more of the encoder unfreezes: from $5{\times}10^{-3}$ and $3{\times}10^{-4}$ in Scenario~A to $5{\times}10^{-4}$ and $5{\times}10^{-5}$ in Scenario~C, where the encoder top, middle, and bottom thirds receive $5{\times}10^{-6}$, $1{\times}10^{-6}$, and $5{\times}10^{-7}$. Baselines train for 100 epochs at batch size 256 and learning rate $10^{-3}$.

\section{Results}
\label{sec:experiments}

\subsection{Experimental Setup}
\label{sec:experimental_setup}

\textbf{Datasets.}
We evaluate CogAdapt on two public cognitive load datasets, CLARE and CL-Drive. CLARE contains ECG, EDA, EEG, and gaze data from participants performing MATB-II multitasking, with self-reported cognitive load labels collected every 10 seconds~\cite{bhatti2024clare}. CL-Drive contains physiological and eye-tracking data from drivers in a simulator, with workload ratings collected every 10 seconds across nine driving complexity levels~\cite{angkan2024multimodal}. We use the ECG modality from both datasets.

\textbf{Labels.}
Both datasets use a 1--9 self-reported cognitive load scale, binarized into low (1--4) and high (5--9) per our baselines protocol. 
\textbf{Preprocessing.}
Preprocessing follows Section~\ref{sec:input}: 500~Hz resampling, a 0.5--40~Hz bandpass filter, 5-second windows (50\% overlap in training, none in evaluation), and per-window, per-lead z-scoring.

\textbf{Evaluation protocol.}
We report leave-one-subject-out (LOSO) cross-validation as the main evaluation setting, since it tests generalization to subjects not seen during training. We report macro-F1 with 95\% bootstrap confidence intervals (5,000 resamples) and AUROC. Checkpoints are selected by AUROC under stratified K-fold, where every fold contains both classes and a threshold-independent criterion is preferable, and by macro-F1 under LOSO, where single-class subject-level validation folds make AUROC undefined  (macro-F1 stays defined; the absent class contributes~$0$ to the average). Significance is assessed with paired Wilcoxon signed-rank tests on per-subject macro-F1, with Holm correction across the six CogAdapt-C--baseline comparisons. We also compute silhouette coefficients on pooled CogAdapt-C embeddings using two label sets 1-participantID and 2-cognitive-load label.

\textbf{Baselines.}
We compare CogAdapt against ECG-LightCNN, a Transformer encoder trained from scratch and an HRV Random Forest (HRV-RF) over 35 time- and frequency-domain features  following~\cite{bhatti2024clare}. All three share CogAdapt's preprocessing, labels, and splits. We re-implement the published baselines because some CLARE folds were not released, and HRV-RF serves as a representative classical algorithm.

\textbf{Implementation.}
All experiments ran on a single NVIDIA V100 GPU. We will release code upon acceptance.

\subsection{Main Results}
\label{sec:main_results}

\begin{table*}[t]
    \vspace{-1em}
\caption{Performance on CLARE and CL-Drive under K-fold and LOSO. Cells are mean\,$\pm$\,std over folds (K-fold 10; LOSO 20 on CLARE, 21 on CL-Drive). AUC is AUROC. Best per column in \textbf{bold}.}
\label{tab:main_results}
\centering
\footnotesize
\renewcommand{\arraystretch}{1.05}
\setlength{\tabcolsep}{3pt}
\resizebox{\textwidth}{!}{%
\begin{tabular}{@{}lcccccccccccc@{}}
\toprule
\multirow{3}{*}{\textbf{Method}} & \multicolumn{6}{c}{\textsc{CLARE}} & \multicolumn{6}{c}{\textsc{CL-Drive}} \\
\cmidrule(lr){2-7} \cmidrule(lr){8-13}
 & \multicolumn{3}{c}{\textbf{K-Fold}} & \multicolumn{3}{c}{\textbf{LOSO}} & \multicolumn{3}{c}{\textbf{K-Fold}} & \multicolumn{3}{c}{\textbf{LOSO}} \\
\cmidrule(lr){2-4} \cmidrule(lr){5-7} \cmidrule(lr){8-10} \cmidrule(lr){11-13}
 & \textbf{Acc\,$\pm$\,std} & \textbf{F1\,$\pm$\,std} & \textbf{AUC} & \textbf{Acc\,$\pm$\,std} & \textbf{F1\,$\pm$\,std} & \textbf{AUC} & \textbf{Acc\,$\pm$\,std} & \textbf{F1\,$\pm$\,std} & \textbf{AUC} & \textbf{Acc\,$\pm$\,std} & \textbf{F1\,$\pm$\,std} & \textbf{AUC} \\
\midrule
ECG-LightCNN~\cite{bhatti2024clare} & $.774{\pm}.007$ & $.726{\pm}.009$ & .832 & $.735{\pm}.136$ & $.514{\pm}.044$ & .539 & $.814{\pm}.017$ & $.808{\pm}.018$ & .887 & $.716{\pm}.114$ & $.607{\pm}.093$ & .707 \\
Transformer~\cite{bhatti2024clare}  & $.706{\pm}.012$ & $.643{\pm}.011$ & .718 & $.640{\pm}.116$ & $.496{\pm}.044$ & .506 & $.688{\pm}.014$ & $.680{\pm}.013$ & .731 & $.617{\pm}.102$ & $.541{\pm}.086$ & .604 \\
HRV-RF ~\cite{bhatti2024clare}      & $.756{\pm}.011$ & $.688{\pm}.016$ & .795 & $.668{\pm}.167$ & $.444{\pm}.065$ & .517 & $.768{\pm}.013$ & $.758{\pm}.014$ & .835 & $.594{\pm}.137$ & $.533{\pm}.123$ & .672 \\
\midrule
CogAdapt-A & $.742{\pm}.012$ & $.717{\pm}.010$ & .843 & $.681{\pm}.122$ & $.527{\pm}.052$ & .598 & $.823{\pm}.018$ & $.821{\pm}.018$ & .906 & $.665{\pm}.166$ & $.578{\pm}.149$ & .705 \\
CogAdapt-B & $.762{\pm}.012$ & $.736{\pm}.012$ & .857 & $.691{\pm}.127$ & $.557{\pm}.085$ & .706 & $.833{\pm}.019$ & $.831{\pm}.019$ & .919 & $.770{\pm}.073$ & $.704{\pm}.099$ & .811 \\
\textbf{CogAdapt-C} & $\mathbf{.813{\pm}.033}$ & $\mathbf{.785{\pm}.034}$ & \textbf{.898} & $\mathbf{.736{\pm}.104}$ & $\mathbf{.626{\pm}.119}$ & \textbf{.799} & $\mathbf{.862{\pm}.025}$ & $\mathbf{.860{\pm}.025}$ & \textbf{.945} & $\mathbf{.831{\pm}.091}$ & $\mathbf{.768{\pm}.117}$ & \textbf{.889} \\
\bottomrule
\end{tabular}}
\end{table*}

Table~\ref{tab:main_results} reports K-fold and LOSO results for all methods on both datasets. Under LOSO, CogAdapt-C reaches macro-F1 0.626 [0.57, 0.68] on CLARE and 0.768 [0.72, 0.81] on CL-Drive, gains of $+0.112$ and $+0.161$ over ECG-LightCNN (95\% bootstrap CIs, 5{,}000 resamples). AUROC rises from 0.539 to 0.799 on CLARE and from 0.707 to 0.889 on CL-Drive. HRV-RF reaches 0.444 on CLARE and 0.533 on CL-Drive. These results suggest that 5-second hand-crafted features do not capture enough cross-subject load information. The large per-fold standard deviation reflects the wide per-subject class balance under LOSO rather than optimization instability, matching prior LOSO results on CLARE~\cite{bhatti2024clare}. On CL-Drive, the participant-ID silhouette is 0.60, while the load-label silhouette is 0.02. Windows from the same participant tend to lie closer together than windows with the same cognitive-load label. This does not contradict the macro-F1 of 0.768. The classification head can still decode load, but the encoder embeddings do not form compact load-based clusters.

Every Wilcoxon signed-rank test across the six CogAdapt-C-vs-baseline comparisons (three baselines, two datasets) remains significant after Holm correction (largest corrected $p < 0.001$; matched-pairs rank-biserial $r$ from $+0.89$ to $+1.00$). CogAdapt-B also beats every baseline under LOSO after correction. The protocol is leak-free: LOSO never exposes the test subject during training, and per-window z-scoring uses only within-window statistics, so no test-subject distribution reaches the model.

\subsection{Effect of Progressive Fine-Tuning}
\label{sec:profine_results}

The scenarios differ only in how much of the encoder is updated. Macro-F1 rises monotonically from A to B to C (Table~\ref{tab:main_results}), so encoder adaptation helps. The frozen encoder (A) trails ECG-LightCNN on driving, but unfreezing the top layers (B) closes that gap, so partial fine-tuning suffices once the domain shifts.

\subsection{LeadBridge Ablation}
\label{sec:leadbridge_ablation}

Table~\ref{tab:leadbridge_ablation} isolates the role of LeadBridge pretraining (architecture in Section~\ref{sec:adapter}). All four mappings share the frozen encoder and run on CLARE/CL-Drive, so any difference between rows comes from the mapping. LeadBridge reaches the highest macro-F1 on both datasets, 0.527 and 0.578, against 0.380--0.409 and 0.508--0.519 for the adapter-free mappings. AUROC tells a more mixed story, since zero-padding and the Dower transform reach higher AUROC than LeadBridge on both datasets. The pretrained mapping therefore helps most with thresholded, class-balanced prediction rather than with ranking discrimination. Without PTB-XL pretraining the random adapter matches the fixed Dower transform, which points to the learned lead relationships, not adapter capacity, as the source of the macro-F1 gain.

\begin{table}[H]
\caption{LeadBridge ablation, all under the \emph{frozen} ECG-FM encoder (Scenario~A); only the $3{\to}12$ mapping and the head are trained. Rows: zero-padding (no adapter), random adapter (untrained Conv1d), Dower (fixed physics transform), and LeadBridge (PTB-XL pretrained). LOSO macro-F1/AUROC; the LeadBridge row equals the CogAdapt-A numbers in Table~\ref{tab:main_results}.}
\label{tab:leadbridge_ablation}
\centering
\small
\renewcommand{\arraystretch}{1.0}
\setlength{\tabcolsep}{4pt}
\begin{tabular}{lcccc}
\toprule
\textbf{3$\to$12 Mapping} & \multicolumn{2}{c}{\textbf{CLARE}} & \multicolumn{2}{c}{\textbf{CL-Drive}} \\
\cmidrule(lr){2-3}\cmidrule(lr){4-5}
 & \textbf{F1} & \textbf{AUC} & \textbf{F1} & \textbf{AUC} \\
\midrule
Zero-padding   & $.380$ & $.639$ & $.519$ & $.764$ \\
Random adapter & $.409$ & $.548$ & $.516$ & $.704$ \\
Dower transform & $.407$ & $.628$ & $.508$ & $.744$ \\
\textbf{LeadBridge} & \textbf{.527} & \textbf{.598} & \textbf{.578} & \textbf{.705} \\
\bottomrule
\end{tabular}
\end{table}

\subsection{LeadBridge Pretraining Quality}
\label{sec:lead_reconstruction}

\begin{figure}[H]
    \centering
    \includegraphics[width=\linewidth]{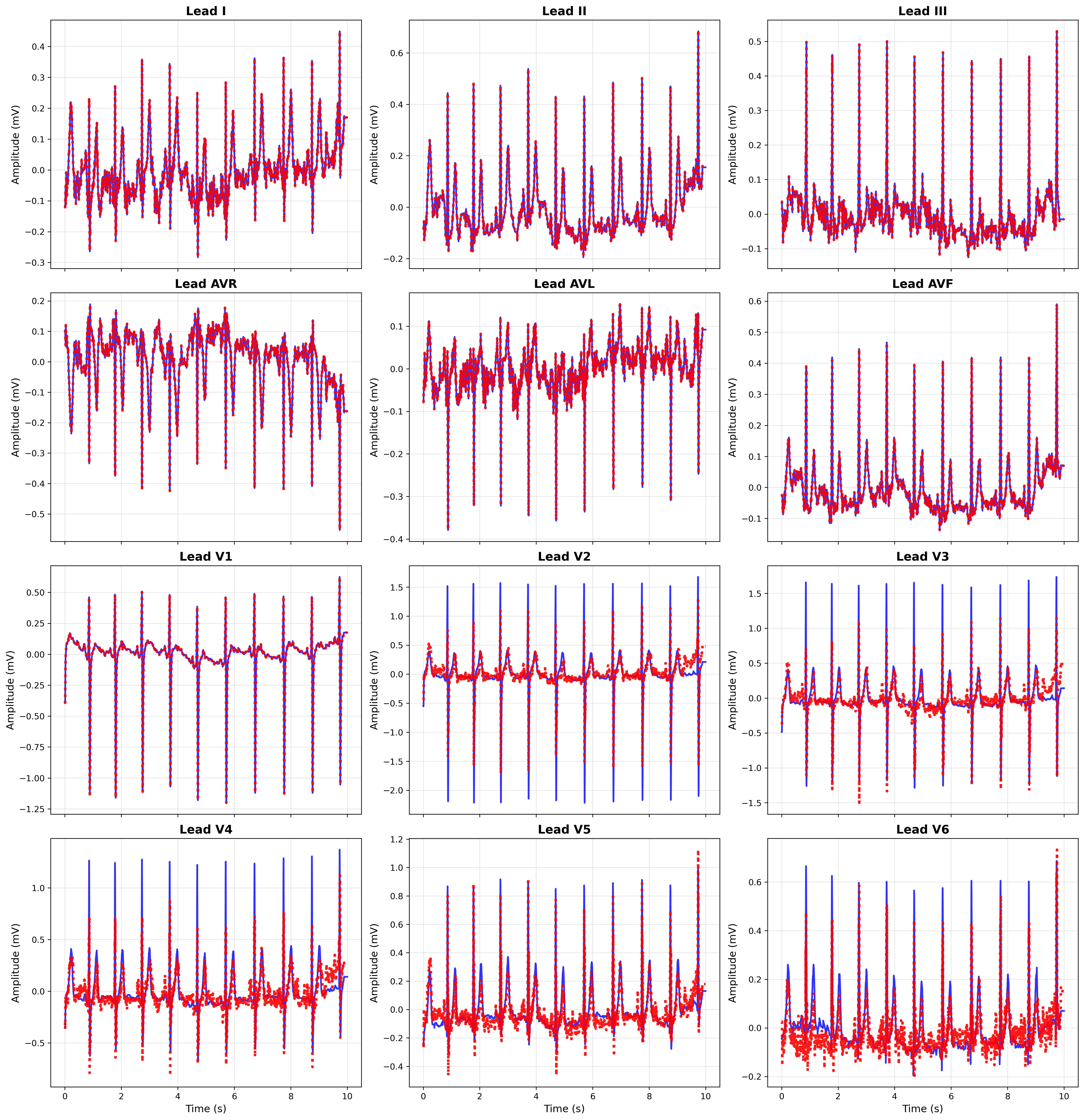}
    \caption{LeadBridge reconstruction on a held-out PTB-XL recording (blue: ground truth, red dashed: output). The limb/augmented leads and the passthrough V1 overlap closely; the learned leads V2--V6 deviate more.}
    \label{fig:lead_reconstruction}
\end{figure}

This analysis checks LeadBridge pretraining on PTB-XL. It does not test wearable transfer. Figure~\ref{fig:lead_reconstruction} shows the 12-lead output for one recording, and Table~\ref{tab:leadbridge_performance} compares per-lead RMSE and correlation against the Dower transform~\cite{dower1980deriving} and linear regression~\cite{obianom2025reconstruction} on held-out PTB-XL. LeadBridge leads on every precordial lead except V6. On wearable recordings the inputs are domain-shifted in noise, electrode placement, and amplitude, yet Tables~\ref{tab:main_results} and~\ref{tab:leadbridge_ablation} show the mapping still transfers. Per-window normalization may help because it removes amplitude shifts before LeadBridge.

\begin{table}[H]
    \caption{ECG lead reconstruction on held-out PTB-XL for precordial leads V2--V6. Green marks the best value per lead. Yellow marks a near-tie.}
    \label{tab:leadbridge_performance}
    \centering
    \renewcommand{\arraystretch}{1}
    \begin{tabular}{l|ccc}
    \hline
    \rowcolor{gray!20}
    \textbf{Lead} & \textbf{Dower} & \textbf{Linear Regression} & \textbf{LeadBridge} \\
    \hline
    \multicolumn{4}{l}{\cellcolor{blue!10}\textsc{RMSE} $\downarrow$} \\
    \textbf{V2} & 205.69 & 195.08 & \cellcolor{green!20}\textbf{175.61} \\
    \textbf{V3} & 290.39 & 194.52 & \cellcolor{green!20}\textbf{189.77} \\
    \textbf{V4} & 270.63 & \cellcolor{yellow!30}\textbf{164.72} & \cellcolor{green!20}\textbf{164.02} \\
    \textbf{V5} & 215.06 & 143.45 & \cellcolor{green!20}\textbf{140.76} \\
    \textbf{V6} & 161.40 & \cellcolor{green!20}\textbf{129.66} & 135.22 \\
    \hline
    \multicolumn{4}{l}{\cellcolor{orange!10}\textsc{Correlation Coefficient (CC)} $\uparrow$ } \\
    \textbf{V2} & 0.546 & 0.546 & \cellcolor{green!20}\textbf{0.669} \\
    \textbf{V3} & 0.569 & 0.570 & \cellcolor{green!20}\textbf{0.613} \\
    \textbf{V4} & 0.662 & 0.698 & \cellcolor{green!20}\textbf{0.718} \\
    \textbf{V5} & 0.679 & 0.694 & \cellcolor{green!20}\textbf{0.717} \\
    \textbf{V6} & 0.544 & 0.549 & 0.547 \\
    \hline
    \end{tabular}
    \end{table}

\subsection{Physiological Validity of Labels}
\label{sec:hrv_validation}

ECG-based load classification assumes that cognitive load modulates cardiac physiology in these datasets. As a descriptive check, we compared HRV features between low- and high-load windows and read the effect sizes (Cohen's $d$); we omit $p$-values because windows within a subject are correlated, which makes a window-level test overstate significance.

On CL-Drive, mean HR ($d = 0.71$), mean RR ($d = 0.65$), and pNN50 ($d = 0.34$) separate the load classes. On CLARE the same features show much smaller effects (mean RR $d = 0.21$, pNN50 $d = 0.19$), and RMSSD and SDNN show essentially none. The autonomic signal is therefore weaker on CLARE, which matches its higher baseline difficulty.

\subsection{Cross-Dataset Transfer}
\label{sec:transfer}

We also test zero-shot transfer between datasets: train on one, test directly on the other (Table~\ref{tab:cross_transfer}). A CLARE-trained model scores F1 0.589 on CL-Drive, down from 0.626 in-domain; a CL-Drive-trained model scores 0.451 on CLARE, down from 0.768. ECG-LightCNN degrades by a similar margin. These drops set a clear boundary on generalization, which we discuss in Section~\ref{sec:discussion}.

\begin{table}[H]
\caption{Zero-shot cross-dataset transfer: train on one dataset, test directly on the other.}
\label{tab:cross_transfer}
\centering
\small
\setlength{\tabcolsep}{4pt}
\begin{tabular}{lcccc}
\toprule
\textbf{Transfer direction} & \multicolumn{2}{c}{\textbf{CogAdapt-C}} & \multicolumn{2}{c}{\textbf{ECG-LightCNN}} \\
\cmidrule(lr){2-3}\cmidrule(lr){4-5}
 & F1 & AUC & F1 & AUC \\
\midrule
CLARE $\rightarrow$ CL-Drive & .589 & .619 & .510 & .522 \\
CL-Drive $\rightarrow$ CLARE & .451 & .483 & .386 & .492 \\
\bottomrule
\end{tabular}
\end{table}

\subsection{Inference Latency}
\label{sec:latency}

Scenarios A, B, and C share one inference graph and differ only in which weights receive gradients during training, so latency is identical across them. On the V100, CogAdapt processes one 5-second window in $18.7\pm1.6$~ms (53~FPS), against $1.28$~ms for the from-scratch baselines (500 runs, batch size~1). Both run far inside the 10-second labeling interval.

\begin{table}[H]
    \caption{Single-sample GPU inference latency (mean $\pm$ std, 500 runs, batch size~1).}
    \label{tab:latency}
    \centering
    \small
    \renewcommand{\arraystretch}{1.0}
    \setlength{\tabcolsep}{4pt}
    \begin{tabular}{lrrr}
    \toprule
    \textbf{Method} & \textbf{Latency (ms)} & \textbf{FPS} \\
    \midrule
    ECG-LightCNN & $1.28 \pm 0.07$ & 784 \\
    Transformer  & $1.28 \pm 0.18$ & 779 \\
    CogAdapt-C   & $18.7 \pm 1.6$  &  53 \\
    \bottomrule
    \end{tabular}
    \end{table}
\section{Discussion and Limitations}
\label{sec:discussion}

CogAdapt-C partial improvement over ECG-LightCNN is consistent with the feature quality of ECG-FM pretrained on 1.5 million ECG recordings versus one trained from scratch on 21 subjects per fold at most. The frozen encoder (Scenario A) matches ECG-LightCNN on CLARE but trails on CL-Drive. So, lead adaptation alone is not enough after the domain shifts, and progressive adaptation supplies the rest. The larger CL-Drive gain matches its larger HRV effect size. Mean RR gives Cohen's $d=0.65$ on CL-Drive and $d=0.21$ on CLARE, so load is more legible there. A clinical ECG foundation model therefore carries structure useful beyond pathology. CLEF and NormWear ~\cite{normwear2025,clef2025} address wearable biosignal pretraining, but they do not evaluate wearable ECG cognitive-load classification. CogAdapt addresses the lead mismatch and the task mismatch in the same using the LeadBridge Adapter and Profine as a practical alternative to training from scratch on small labeled datasets. ProFine also preserves the pretrained backbone: centered kernel alignment between pretrained and fine-tuned encoders averages 0.894 on CLARE and 0.840 on CL-Drive, with relative $\ell_2$ weight drift below 0.35\% on both datasets. These values suggest that ProFine changes the encoder without erasing its pretrained between subject structures.

Two limits matter for deployment. CogAdapt-C ranks load well but overestimates its own confidence: on CL-Drive the expected calibration error is 0.175, more than double that of ECG-LightCNN (0.076). Therefore, probability-sensitive use should apply temperature scaling first. Cross-dataset transfer drops without retraining (Section~\ref{sec:transfer}), in line with other physiological classifiers~\cite{gjoreski2020cogload}: the learned representations are dataset-specific, and few-shot cross-environment adaptation, evaluated on a larger multi-site cohort, is the next step.

\subsection*{Limitations}

CLARE and CL-Drive include 20 and 21 subjects. Larger and more diverse cohorts need separate validation. Self-reported labels include rating bias and delay. Binary labels simplify a continuous construct. Ordinal or regression targets may capture smaller load changes. LeadBridge produces a 12-lead-compatible input for the encoder, not a clinical-quality reconstruction. We evaluate offline. Streaming use still needs testing, although one window takes 18.7 ms on a V100. Finally, cross-dataset transfer degrades without retraining, which bounds generalization.

\section{Conclusion}
\label{sec:conclusion}

We presented CogAdapt for adapting a clinical ECG foundation model to wearable cognitive-load assessment. LeadBridge, a light adapter pretrained on PTB-XL, closes the 3-lead-to-12-lead sensor gap. ProFine adapts the encoder with staged unfreezing and bucketed layer-wise learning rates and closes the task gap. Under leave-one-subject-out evaluation, CogAdapt reaches macro-F1 of 0.626 on CLARE and 0.768 on CL-Drive, gains of $+0.112$ and $+0.161$ over the best from-scratch baseline, both significant by a Wilcoxon signed-rank test. Descriptive HRV effect sizes indicate the targeted autonomic signal in both datasets, and inference runs in 18.7~ms per window. Cross-dataset transfer still drops without retraining. Future work should target cognitive-load representations that transfer across environments.

{\footnotesize
\bibliographystyle{IEEEtran}
\bibliography{ijcai26}}

\end{document}